\pdfoutput=1
\documentclass[a4paper,twoside,11pt]{article}
\usepackage{a4wide,fancyhdr}
\usepackage{hyperref}
\hypersetup{
    colorlinks=true,
    linkcolor=black,
    citecolor=black,
    filecolor=black,
    urlcolor=black,
}

\usepackage{microtype}
\usepackage{graphicx}
\usepackage{subfigure}
\usepackage{booktabs} 
\usepackage{algorithm}
\usepackage{algorithmic}
\usepackage{amsmath}

\title{\vspace{-30pt}Network of Evolvable Neural Units: \\ Evolving to Learn at a Synaptic Level}
\author{Paul Bertens\footnote{Github: {https://github.com/paulbertens}, e-mail: {p.m.w.a.bertens@gmail.com}} $^{,1}$, Seong-Whan Lee$^{1,2}$ \\ 
\noindent $^{1}$Department of Brain and Cognitive Engineering, \\
\noindent $^{2}$Department of Artificial Intelligence,\\ Korea University, Seoul, South Korea
}

\begin{document}

\maketitle
\begin{abstract}
Although Deep Neural Networks have seen great success in recent years through various changes in overall architectures and optimization strategies, their fundamental underlying design remains largely unchanged. Computational neuroscience on the other hand provides more biologically realistic models of neural processing mechanisms, but they are still high level abstractions of the actual experimentally observed behaviour. Here a model is proposed that bridges Neuroscience, Machine Learning and Evolutionary Algorithms to evolve individual soma and synaptic compartment models of neurons in a scalable manner. Instead of attempting to manually derive models for all the observed complexity and diversity in neural processing, we propose an Evolvable Neural Unit (ENU) that can approximate the function of each individual neuron and synapse. We demonstrate that this type of unit can be evolved to mimic Integrate-And-Fire neurons and synaptic Spike-Timing-Dependent Plasticity. Additionally, by constructing a new type of neural network where each synapse and neuron is such an evolvable neural unit, we show it is possible to evolve an agent capable of learning to solve a T-maze environment task. This network independently discovers spiking dynamics and reinforcement type learning rules, opening up a new path towards biologically inspired artificial intelligence.
\end{abstract}

\section{Introduction}

Much research has been done to understand how neural processing works \cite{hassabis2017neuroscience, spruston2008pyramidal, sacramento2018dendritic}, including synaptic learning \cite{abbott2000synaptic} and overall network dynamics \cite{borgers2003synchronization, fell2011role}. However these processes are extremely complex and interact in a way that is currently still not well understood. Mathematical models are commonly used to approximate neurons, synapses and learning mechanisms \cite{abbott1999lapicque, abbott2000synaptic}. However, neurons can behave in many different ways, and no single model can accurately capture all experimentally observed behaviour. The neural complexity is mostly at the small scale, i.e. the cellular local interactions that can lead to an intelligent overall system. While information processing at the network level is generally well understood (and the basis for artificial neural networks \cite{jain1996artificial, lecun2015deep}), it’s how individual neurons are actually capable of giving rise to the overall networks behaviour and learning capability that is largely unknown.

\begin{figure}[htb]
\centering
\includegraphics[width=0.6\columnwidth]{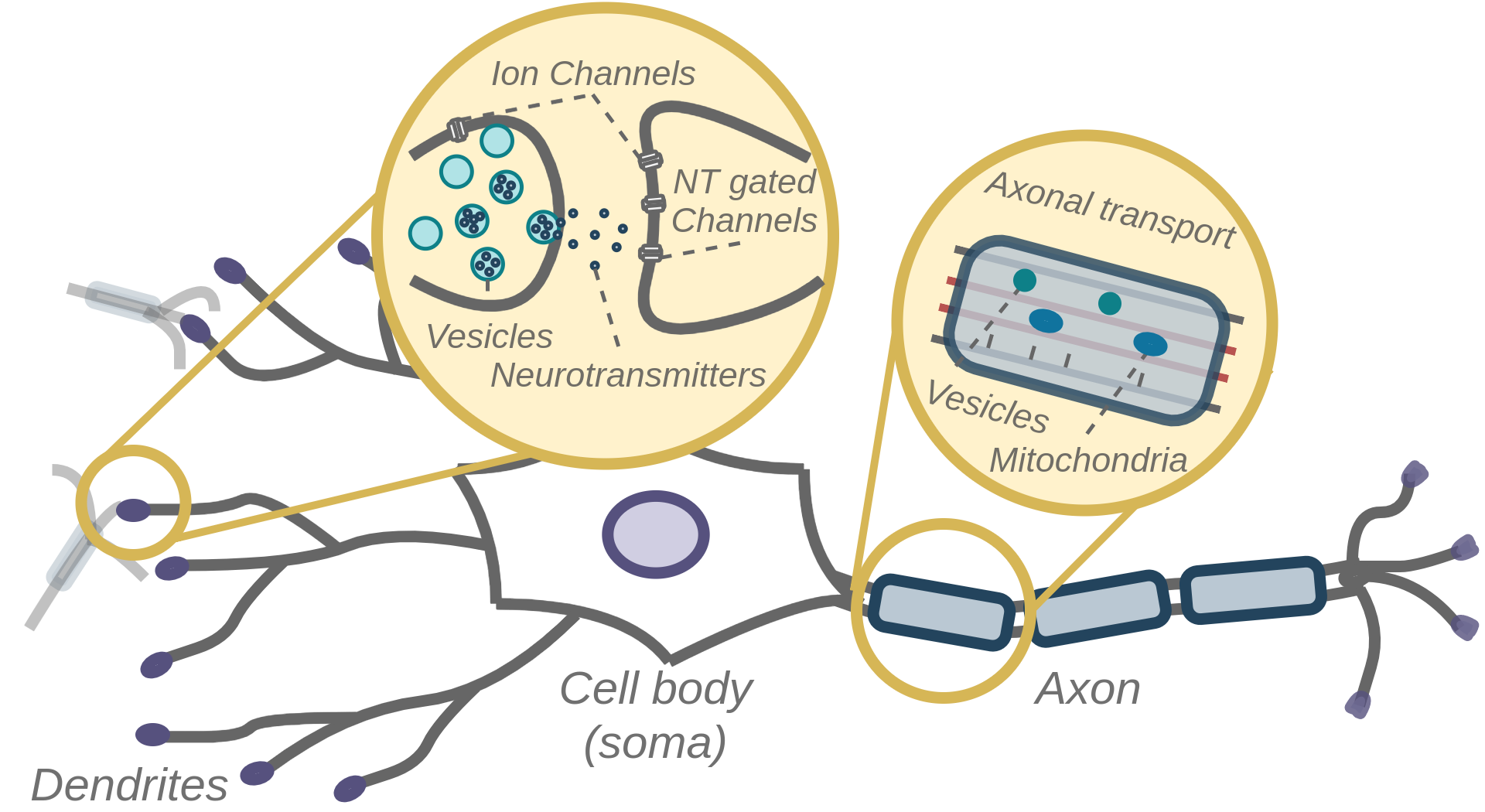}
\caption{Abstract overview of a single neuron. Functionally the dendrites receive information, the cell body integrates and transforms that information and the axon transmits it to other neurons. Neurotransmitters allow for communication between neurons, and the number of neurotransmitters released affect the strength of the synaptic connection. Learning then occurs by updating this synaptic strength depending on the information received. Besides graded and action potentials (spikes), complex transport of mitochondria, vesicles used to contain neurotransmitters and other information that can change the behaviour of the neuron also occur along the axon and dendrite.}
\vspace{-0.25cm}
\label{neuron_synapse_detail}
\end{figure}

\paragraph{Biological Neural Networks}
When looking in more detail at the smaller scale of neurons and synapses, complexity quickly arises. Different ion-channels in the neuron are responsible for generating action potentials (spikes) or graded potentials (real valued), mainly sodium ($Na^+$), potassium ($K^+$) and calcium ($Ca^{2+}$) channels \cite{catterall1995structure}. These ion channels are also responsible for triggering the release of chemical neurotransmitters, the main method of communication between neurons in biological neural networks. Many type of neurotransmitters exist that each have distinct roles, e.g. Dopamine \cite{flagel2011selective} (related to learning), GABA\textsubscript{B} \cite{mccormick1989gaba} (inhibitory), Acetylcholine \cite{levitan2015neuron} (motor neurons) and Glutamate \cite{nedergaard2002beyond} (excitatory). Morphological differences across the brain are also common, where different type of neurons exist in different regions of the brain and across different cortical layers \cite{mountcastle1997columnar}. 

Axons and dendrites actively transport other information \cite{vale2003molecular}, for example vesicles \cite{hilfiker1999synapsins} (which hold neurotransmitters) and mitochondria \cite{hollenbeck2005axonal} (required for membrane excitability and neuroplasticity). Many protein types are also common \cite{vale2003molecular}, which are responsible for a diver set of functions, including increasing or decreasing the number of neuroreceptors in the synapses \cite{collingridge2004receptor}, and aiding in the release of neurotransmitters \cite{hilfiker1999synapsins}. Furthermore, neurons can exhibit both regular or bursting spiking behaviour depending on their internal state and neuronal type \cite{kepecs2002bursting, kepecs2003information}. Deriving models capable of capturing all this behaviour is thus an active research area. These mechanisms might however simply be the result of biological constraints and evolutionary processes, and it is still uncertain which structures exactly are required to exist for intelligent behaviour to emerge.

Generally models in the computational neuroscience field focus on modeling individual neurons and synaptic learning behaviour. Simple Integrate and fire neurons \cite{abbott1999lapicque} (IAF) assume input potential is summated over time, and once this reaches some threshold a spike occurs. More advanced models also try to model the sodium and potassium ion channels as done in the Hodgkin–Huxley (HH) model \cite{hodgkin1952quantitative}, which gives more realistic action potentials at the output.

Hebbian plasticity was one of the first proposed models describing synaptic learning  \cite{hebb1949organization}, briefly it states that the synaptic weight increases if the pre-synaptic neuron fires before the post-synaptic neuron. This change in synaptic weight can then increase or decrease the firing rate of the post-synaptic neuron in the future when receiving a similar stimuli. A more extensive model that also takes the spike timing into account is Spike Timing Dependent Plasticity \cite{abbott2000synaptic} (STDP), which updates the weights depending on the time between the pre and post synaptic neuron.

\paragraph{Artificial and Recurrent Neural Networks}
Deep Learning models have been very successful in many practical applications \cite{lecun2015deep}, and although spiking neural networks (SNNs) have been developed \cite{maass1997networks, bellec2018long}, so far they have seen limited success due to their high computational demand and difficulty in training. Artificial Neural Networks, the foundation of deep learning models, are especially effective in performing function approximation, and are known to be universal function approximators \cite{hornik1991approximation}. Recurrent Neural Network (RNNs) architectures have also been investigated in order to process sequential data and allow memory to be stored between successive time steps \cite{lecun2015deep}, most commonly Long-Short Term Memory \cite{hochreiter1997long} (LSTMs) and Gated Recurrent Units \cite{cho2014learning} (GRUs). These add gating mechanisms to standard RNNs to allow for easier learning of long-range dependencies and to avoid vanishing gradients in back-propagation. 

Deep learning based models however suffer from several limitations \cite{hassabis2017neuroscience}. They require backpropagation of errors through the whole network, have difficulty performing one-shot learning (learning from a single example) and suffer from catastrophic forgetting of a previous task once trained on a new task. They are also extreme abstractions of biological neural networks, only considering a single weight value on the connections and having a single real valued number as their output.

\paragraph{Evolutionary Algorithms}
Evolutionary algorithms have been widely applied to a variety of domains \cite{back1996evolutionary, stanley2019designing}, and many related optimization algorithms exist that are based on the evolutionary principles of mutation, reproduction and survival of the fittest. Recently Evolution Strategies (ES) have been successfully applied to train deep neural networks, despite the large amount of parameters of most neural network architectures \cite{salimans2017evolution}. The advantage of ES is that it allows for non-differentiable objective functions, and when training RNNs does not require backpropagation through time. This means we can optimize and evolve the parameters of a model to solve any desired objective we want, and potentially learn over arbitrarily long sequences. 

\section{Objective}
Past attempts on modeling biological neural networks have mostly been focused on manually deriving mathematical rules and abstractions based on experimental data, however in this paper a different approach is taken. The functionality of different components in neural networks are approximated using artificial Recurrent Neural Networks (RNNs), such that each synapse and neuron in the network is a recurrent neural network. To make this computationally feasible the RNN weight parameters across the global network are shared, while each RNN keeps their own internal state. This way the number of parameters to be learned are significantly reduced, and we can use evolutionary strategies to train such a network. Due to the universal function approximation properties of RNNs, there is essentially no limitations on the neuronal behaviour we can obtain.

The RNN memory can store dynamic parameters that determine the behaviour of our compartments (e.g. synapse or neural cell body), while the shared weights are evolved. This is related to meta-learning approaches \cite{schmidhuber1987evolutionary, bengio1992optimization, schmidhuber1992learning, andrychowicz2016learning, ha2016hypernetworks}, where we are learning to learn. Since all RNN weights are shared across all synaptic or neural compartments, learning different behaviour can only occur by learning (evolving) to store and update dynamic parameters in the RNN memory state.

Such an RNN can thus be seen as modeling a function that expresses the neuronal behaviour. However, instead of having a fixed mathematical function derived from experimental observation (like e.g. hebbian learning, integrate-and-fire neurons or STDP rules), we evolve a function that could exhibit much more diverse behaviour. In our case it is not required that the evolved behaviour has to match experimental data or preexisting models, as cells in the real world are bound by more physical constraints than in our simulated setting. Ultimately we would like to evolve a type of mini-agent that when duplicated and connected together in an overall neural network exhibit intelligent learning behaviour. To this end, network agents containing these mini-agents are evolved using evolutionary algorithms, where their fitness is their ability to learn new tasks.

\subsection{Contributions}
To summarize, we are thus attempting to evolve a new type of Neural Network where each individual neuron and synapse in the network is by itself modeled by an evolvable Recurrent Neural Network. Our main contributions are as follows: 
\begin{itemize}
\setlength\itemsep{1pt}
\item Propose an Evolvable Neural Unit (ENU) that can be evolved through Evolution Strategies (ES) which is able to learn to store relevant information in its internal state memory and perform complex processing on the received input using that memory.
\item Demonstrate such an ENU can be evolved to approximate the neural dynamics of simple Integrate and Fire neurons (IAF) and synaptic Spike-Timing-Dependent Plasticity (STDP). 
\item Combine multiple ENUs in a larger network, where the weight parameters of the ENUs to be evolved are shared across the neurons and synaptic compartments, but each internal state is unique and updated based on the local information they receive. 
\item Efficiently evolve and compute such a network of ENUs through large scale matrix multiplications on a single Graphical Processing Unit (GPU). 
\item Show it is possible to evolve a network of ENUs capable of reward based reinforcement learning purely through local dynamics, i.e. evolving to learn.
\end{itemize}

\section{Proposed Model}
\subsection{Evolvable Neural Units (ENUs)} 
As the basis for an evolvable Recurrent Neural Network (RNN) to approximate neural and synaptic behaviour we build upon previous work on modeling long-term dependencies through Gated Recurrent Units \cite{cho2014learning} (GRUs). We extend upon this model and add an additional output gate that applies a non-linear activation function and feeds this back to the input, which simultaneously also reduces the number of possible output channels (improving computational complexity). Additionally they are implemented in such a way that allows them to be combined and evolved efficiently in a larger overall network. We term these units Evolvable Neural Units (ENUs).

\begin{figure}[htb]
\centering
\includegraphics[width=0.85\columnwidth]{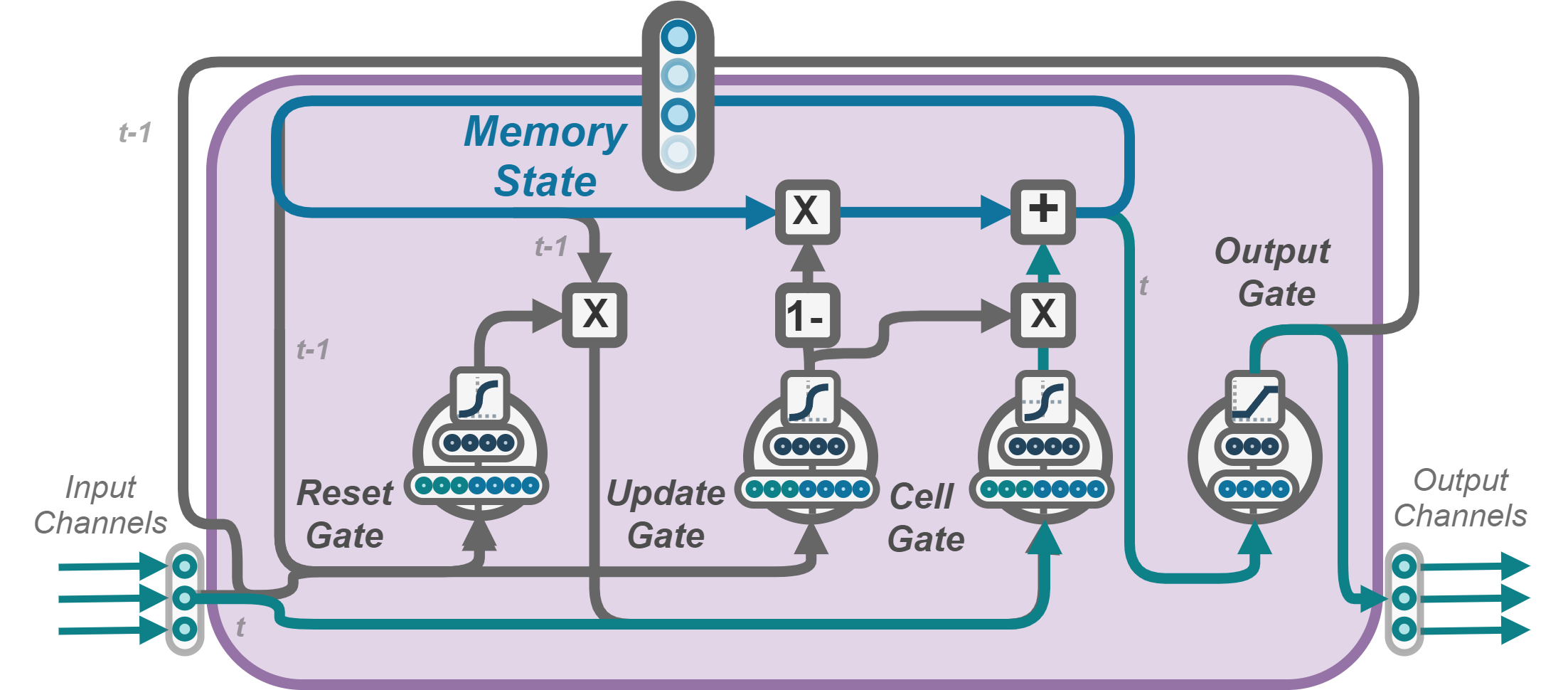} 
\caption{\textbf{An Evolvable Neural Unit}. Different evolvable gating mechanisms protect the internal memory of the neural unit (storing "dynamic parameters"). This memory can potentially encode and store the neuronal behaviour and determines how information is processed (analogous to e.g. the membrane potential, synaptic weights, protein states etc.). The reset gate can forget past information, the update gate determines how much we change the memory state (our dynamic parameters) and the cell gate determines the new value of the dynamic parameters. An additional output gate is also used to reduce the number of output channels and to allow for spike generation to potentially evolve. In this diagram example we have 4 dynamic parameters, 3 input channels and 3 output channels.} 
\label{diagram-enu-detail}
\end{figure}

Using a gating structure allows us to have fine grained control over how different input influences the internal memory state (storing "dynamic parameters"), which in turn controls how that input is processed and how the dynamic parameters are updated. The input and output of the ENU is a vector, where each value in the vector can be considered a type of "channel" used to transmit information between different ENUs. The internal memory state is also a vector, that can store multiple values. It enables us to condition the received input on the dynamic parameters stored in the memory state, which control both the output and gating behaviour of the ENU (see figure \ref{diagram-enu-detail}). Each gate in the model is a standard single layer artificial neural network with $k$ output units and an evolvable weight matrix $w$, where $k$ is equal to the internal memory state size. These gates can process information from the current input channels and previous memory state. This allows us for example to evolve a function that adds some value to a dynamic parameter in the memory state, but only if there is a spike at a certain input channel. Evolving such units to perform complex functions then becomes significantly easier than in standard RNNs, which can suffer from vanishing values and undesired updates to their internal memory state \cite{hochreiter1997long}. Protecting updates to the memory state is especially important in our case, since they define the dynamic parameters that determine the behaviour of our ENU, and should be able to persist over their entire lifetime.

Intuitively this ENU type architecture allows us for example to evolve spike based behaviour through storing and summing received input into the memory state. Then once some threshold is reached the output gate can evolve to activate and output a value, a "spike". This spike is then fed back into the input allowing the reset and update gate to evolve to reset the internal memory state. 

For synaptic compartments the internal state could evolve to memorize when a pre and post synaptic spike occurs , and also store and update some dynamic "weight" parameter that can change how the input is processed and passed to the post-synaptic neuron. Additionally, multiple input and output channels allow flexibility in evolving different type of information processing mechanisms, e.g. using a type of neurotransmitter, or even functions analogous to dendritic and axonal transport.

\subsection{Network of ENUs} 

\begin{figure}[htb]
\centering
\includegraphics[width=0.9\columnwidth]{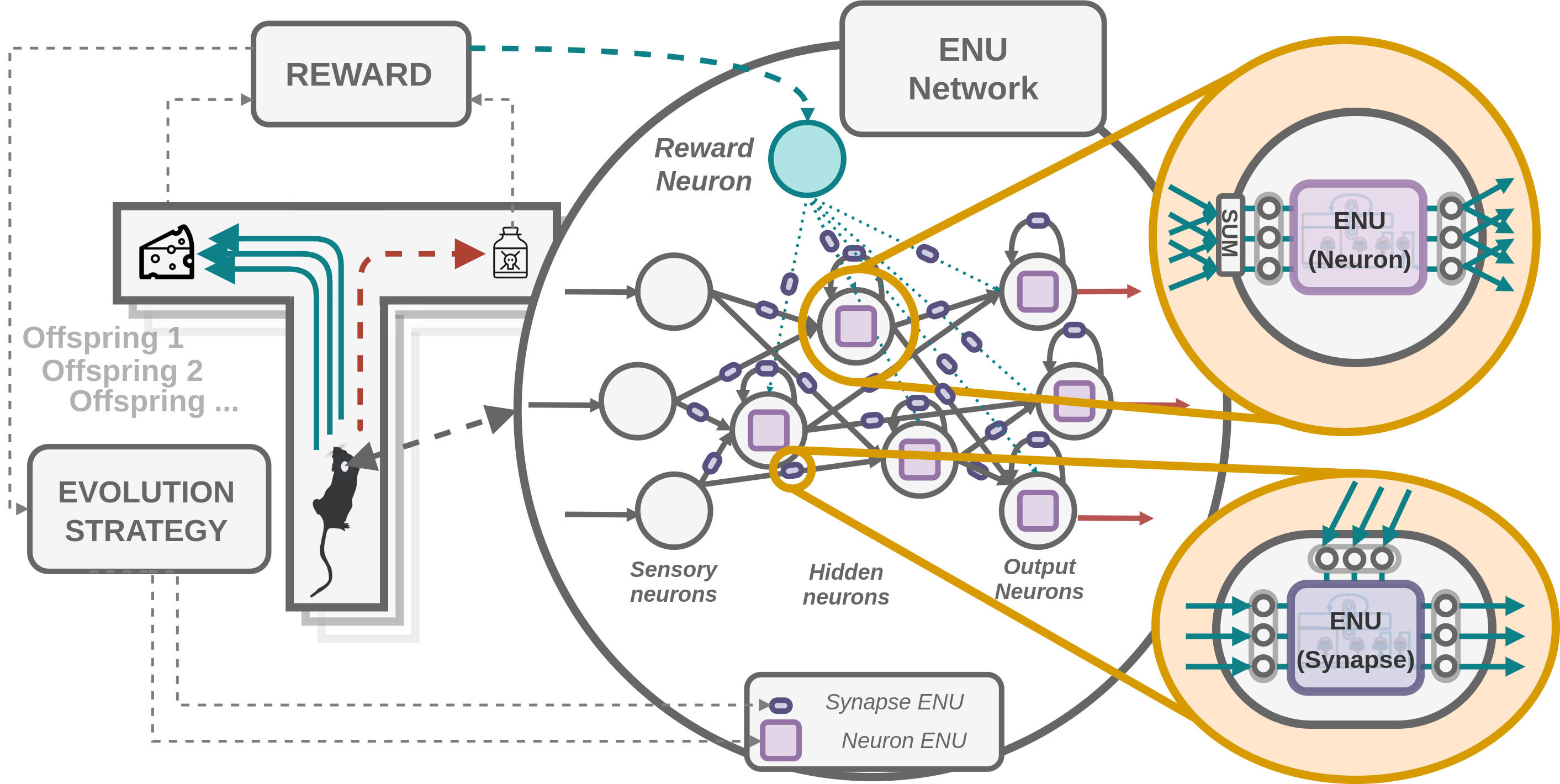}
\caption{\textbf{Network of ENUs}. Example of evolving reinforcement based learning to solve a T-maze environment task. Reward from the environment determines the fitness of the agent. This reward is also passed to the reward neuron in the ENU Network, which connects to the other ENU neurons, allowing them to evolve to update their incoming synapse ENUs according to the reward obtained. Several sensory input neurons are used to detect colors in front of the agent, and the output neurons determine the action the agent has to take (forward, left or right). Each ENU is capable of updating their unique internal dynamic parameters depending on the current and past input, which changes how information is processed over multiple channels and compartments within the overall ENU network.}
\label{experiments_diagram_maze}
\vspace{-0.1cm}
\end{figure}

By combining multiple ENUs in a single network we can construct a ENU based Neural Network (ENU-NN), this network connects multiple neural and synaptic compartment ENU models together as seen in figure \ref{experiments_diagram_maze}. All soma and synaptic neuronal compartments share the same ENU gate parameters, and we only evolve the weights of these shared gates. This means we have two "chromosomes" to evolve, one for the synapses and one for the neurons, shared across all synapses and neurons. However, they each have unique internal memory state variables which allows for complex signal processing and learning behaviour to occur, as the compartments can evolve to update their internal states depending on the local input they receive. Synaptic plasticity (the synaptic weights) in such a model would thus no longer be encoded in the weights of the network directly, but could instead evolve to be stored and dynamically updated in the internal states of the synapse ENUs as a function of the current and past input. 

To evolve reinforcement type learning behaviour in such a network we can construct a sparse recurrent network with several input sensory neurons that detect e.g. different colors in front of the agent (which has an ENU network) in a given environment. We can also designate some ENU neurons as output motor neurons that determine the action the agent should take. Additionally, to allow reward feedback we can have a reward neuron that uses different ENU input channels to indicate environmental rewards obtained (see Figure \ref{experiments_diagram_maze}).

The main components of the ENU network are as follows:
\begin{itemize}
\item \textbf{Neuron ENUs}: Analogous to the biological cell body and axon. Responsible for transforming the summated input from its connected synapses. Can evolve to use multiple channels to transmit information analogous to e.g. spikes or neurotransmitters. Also propogates it's output backwards to all connected input synapses, this way the synapse ENU compartment can potentially learn STDP type learning rules based on post-synaptic neural behaviour. 
\item \textbf{Synapse ENUs}: Analogous to the biological dendrite and synapse. Transforms information from the pre-synaptic to post-synaptic ENU neuron, similar to weights in artificial neural networks. However, it is allowed to use multiple channels to learn to transmit information analogous to e.g. spikes, graded potential or other types of dendritic transport. 
\item \textbf{Integration step}: Sums up the output per channel of each incoming synapse connection, which is then processed by the post-synaptic Neuron ENU. 
\end{itemize}

\section{Method}

\subsection{Evolution Strategies}

Since our desired goal is to solve Reinforcement Learning (RL) based environments and the task is not directly differentiable, we cannot use standard Back-propagation through time (BPTT) to optimize our network. We would also like to be able to learn over long time-spans, which makes BPTT impractical due to vanishing gradient issues. This makes Evolution Strategies (ES) ideally suited \cite{back1991survey, beyer2002evolution, wierstra2014natural, lehman2018more}, and we use a similar approach taken in previous work \cite{salimans2017evolution}. The gradient is approximated through random Gaussian sampling across the parameter space, and the weights are updated following this approximate gradient direction (see Algorithm \ref{algo_es}). We can also use standard Stochastic Gradient Descent (SGD) methods like momentum \cite{sutskever2013importance} on the resulting approximate gradients obtained. Due to weight sharing we only have to evolve the gate parameters of two ENUs, one for the synapse, and one for the neuron, giving us a relatively small parameter space.

\begin{algorithm}[htb]
   \caption{Evolution Strategies}
   \label{alg:example}
\begin{algorithmic}
    \STATE {\bfseries Input:} base parameters $\theta_k$, learning rate $\alpha$, standard deviation $\sigma$
    \FOR{$k=1$ {\bfseries to} $N$}
        \STATE Sample offspring mutations $\epsilon_i,...,\epsilon_n$ from $N(0, I)$
        \STATE Compute fitness $F_i=F(\theta_k + \sigma\epsilon_i)$ for $i = 1,...,n$
        \STATE Calculate approximate gradient $G_k=\sum^n_{i=0} F_i\epsilon_i$
        \STATE Update base parameters $\theta_{k+1}=\theta_k+\alpha G_k$
    \ENDFOR
\end{algorithmic}
\label{algo_es}
\end{algorithm}

\vspace*{-0.3cm}
\paragraph{Fitness Ranking} 
We use fitness ranking in order to reduce the effect of outliers \cite{salimans2017evolution}. We can sort and assign rank values to each offspring according to their fitness, which determines their relative weight in calculating our approximate gradient. We then get a transformed fitness function $F_{r}(\theta_i)=\frac{rank(F(\theta_i))^5}{\sum^n_{i=0} rank(F(\theta_i))^5}$, where rank() assigns a linear ranking from 1.0 to 0.0. This results in around the top 20\% best performing agents to account for 80\% of the gradient estimate. 

\paragraph{Batching} 
Mini-batches were also used to better estimate the fitness of the same offspring across multiple environments, so the fitness of each offspring was determined by the mean fitness across $m$ mini-batch environments. 

\subsection{Gating in an Evolvable Neural Unit}

The exact equations for the ENU gates and updating of the memory state are as follows:
\begin{equation}
\begin{aligned}
    & z_t = \sigma(W_z \cdot [h_{t-1}, o_{t-1}, x_t])\\
    & r_t = \sigma(W_r \cdot [h_{t-1}, o_{t-1}, x_t])\\
    & \widetilde{h_t} = tanh(W_c \cdot [r_t \odot h_{t-1}, o_{t-1}, x_t])\\
    & h_t = (1-z_t) \odot h_{t-1} + z_t \odot \widetilde{h_t}\\
    & o_t = clip((W_o \cdot h_t), 0, 1)\\
\end{aligned}
\end{equation}
Where $x_t$ is the input, $z_t$ the update gate, $r_t$ the reset gate, $\widetilde{h_t}$ the cell gate, $h_t$ the new memory state and $o_t$ the output gate. $\sigma$ is the sigmoid function, '$\cdot$' is matrix multiplication and $\odot$ is element-wise multiplication. $W_*$ are the weight matrices of each gate, and the parameters to be evolved. We can also apply clipping (restricting output values to be between 0 and 1), since we use evolution strategies to optimize our parameters and do not require a strict differentiable activation function. This clipping results in a thresholded output and prevents exploding values. 

\subsection{Implementation}
\begin{figure*}[htb]
\centering
\includegraphics[width=\textwidth]{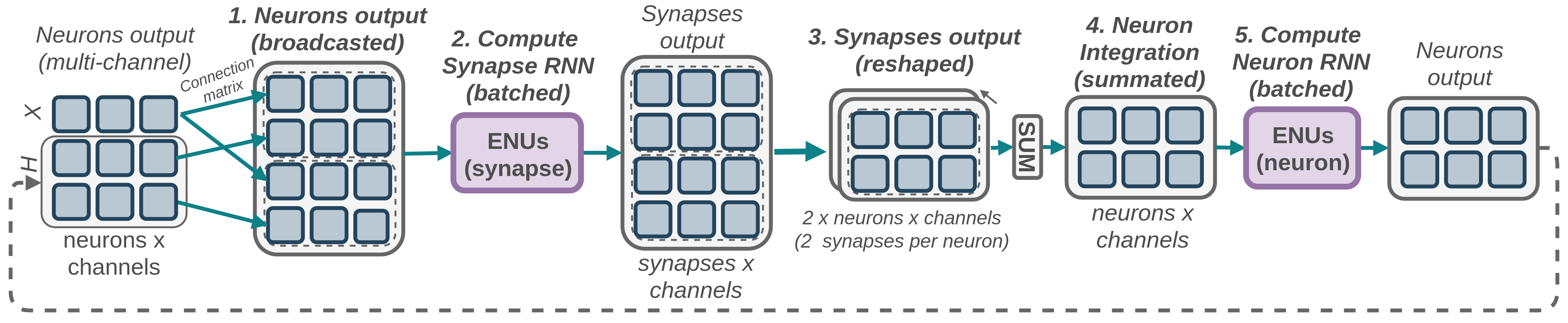} 
\caption{\textbf{Computation flow diagram}. The sensory input neurons X are concatenated with all the ENU neurons H to get our input batch. A connection matrix is then applied that broadcasts (copies) the neurons output to each connected synapse (1). On this resulting matrix we can then apply standard matrix multiplication and compute our synapse ENUs output in parallel (2). We can reshape this and sum along the first axis, as we have the same number of synapses for each neuron (3). This gives us the integrated synaptic input to each neuron (4). Finally, we apply the neuron ENUs on this summated batch and obtain the output for each neuron in the ENU network (5).}
\label{diagram-enu-computation}
\end{figure*}

Figure \ref{diagram-enu-computation} shows a detailed computation diagram, which relies heavily on large scale matrix multiplication. We can reshape each neuron and synaptic compartment output to batches and perform standard matrix multiplication to compute the full ENU-NN network in parallel (since our ENU parameters are shared). The connection matrix determines how neurons are connected, and in this case can be either random or have a fixed sparse topology. It determines how we broadcast the output of a neuron to each synapse connected to it (Broadcasting "copies" the output of neurons to different synapses as multiple synapses can be connected to the same neuron). However, the denser the connection matrix, the more synapses we have and thus the higher our computational cost.

ES can be performed efficiently on a Graphical Processing Unit (GPU) through multi-dimensional matrix multiplications, allowing us to evaluate thousands of mutated networks in parallel. The weight matrix in this case is 3 dimensional, and the 3th dimension stores each offspring's mutated weights. 

Normal matrix multiplication is of the form $(N,K)$ x $(K,M)\rightarrow(N,M)$, where ($N$ is the batch size, $K$ the number of inputs units and $M$ the number of output units), in the 3D case we get $(P,N,K)$ x $(P,K,M)\rightarrow(P,N,M)$, where $P$ is the number of offspring. PyTorch \cite{paszke2017automatic} was used for computing and evolving our ENU-NN, while the rest was implemented mainly in Numpy \cite{oliphant2006guide}, including custom vectorized experimental environments.

\subsection{Experiments}

First we evolve an ENU to mimic the behaviour of a simple Integrate and Fire (IAF) and Spike Timing Dependent Plasticity (STDP) model. We then combine multiple neural and synaptic ENUs into a single network, and evolve reinforcement type learning behaviour purely through local dynamics. 

\begin{figure}[htb]
\centering
\includegraphics[width=\columnwidth]{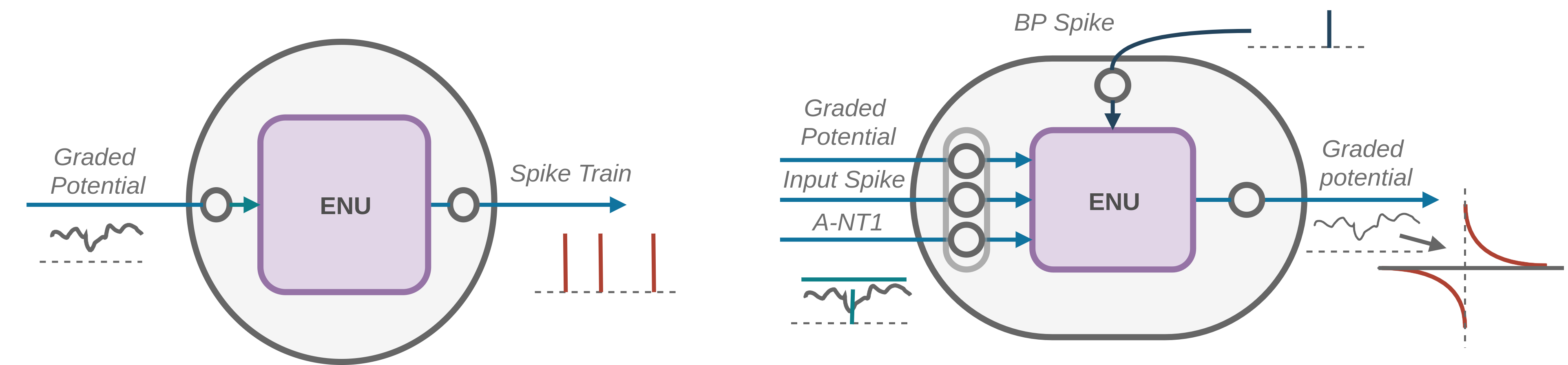} 
\caption{\textbf{IAF and STDP experimental setup}. For evolving the IAF ENU a single random graded potential is given as input, with as goal to mimic the IAF spiking model (left). In case of evolving the STDP rule (right) multiple input channels are used. The graded input potential, the input spike, the neuromodulation signal (A-NT1) and the backpropagating spike. The target is then to output the modified graded input potential matching the STDP rule. }
\label{experiments_diagram}
\end{figure}

\newpage
\paragraph{Evolving Integrate and Fire Neurons (IAF)}
The ENU receives random uniform noise as graded input potential, and has to mimic the IAF model receiving the same input. Once the sum of the input potentials reaches a certain threshold, we reset the internal state of the IAF and output a spike (See also figure \ref{experiments_diagram}). The simplest IAF variant with linear summation is give below:

\begin{equation}
  IAF(x, t, h)=\begin{cases}
    0; \hspace{5pt} $$h_{t} = h_{t-1} + x$$, & \text{if $h<th$}\\
    1; \hspace{5pt} $h=0$, & \text{if $h>=th$}
  \end{cases}
 \label{eq_iaf}
\end{equation}
Where $x$ is the input potential, $t$ the current time, $h$ the membrane potential and $th$ the threshold.

Learning spike like behaviour can generally be difficult since the task is not directly differentiable. Optimizing the mean squared error between the IAF model and ENU output directly is infeasible since even a small shift in spikes would cause a decremental effect on the error. However, since we are using Evolution Strategies we can have more flexible loss functions and optimize the timing and intensity of each spike. Therefore, we optimize the Inter-Spike Interval (ISI) instead, which allows for incremental fitness improvements and gradually gets each spike to better match the desired timing of the IAF model. We also add an additional term that requires each spike to be as close as 1 as possible (matching the IAF spike). 

\paragraph{Evolving Spike-timing dependent plasticity (STDP)}
For evolving a STDP type learning rule we require multiple input channels for the model. The basic equation for the STDP rule is as follows: 
\begin{equation}
  W(t)=\begin{cases}
    $$Ae^{(-\frac{t}{\tau})}$$, & \text{if $t>0$}\\
    $$-Ae^{(\frac{t}{\tau})}$$, & \text{if $t<0$}
  \end{cases}
 \label{eq_stdp}
\end{equation}
where $A$ and $\tau$ are constants that determine the shape of the STDP function, t is the relative timing difference between the pre and post synaptic spike and $W$ is the resulting synaptic weight value.  

First we input random uniform noise as graded potential, we then also generate a random input spike from the pre-synaptic neuron at some time $t_i$ and a random backpropagating spike from the post-synaptic neuron at $t_b$. We also use a neuromodulated variant of the standard STDP rule, which is known to play a role in synaptic learning \cite{pawlak2010timing, fremaux2016neuromodulated}, and only allows STDP type updates to occur if a given input neurotransmitter is present (See figure \ref{experiments_diagram}).

In total we thus have 4 input channels, and a single output channel that is the transformed graded input potential multiplied by some weight value $w$. The weight value is updated according to the standard STDP rule which is dependent on the timing between the pre and post synaptic spike. The fitness is then the mean squared error between the desired STDP model output and ENU output.

\paragraph{Evolving Reinforcement Learning in a network of ENUs}
In order to evolve reinforcement type learning behaviour, we design an experimentally commonly used maze task \cite{deacon2006t} (see figure \ref{experiments_diagram_maze}). The goal of the agent (i.e. the mouse), is to explore the maze and find food, once the agent eats the food he will receive a positive reward and be reset to the initial starting location. On the other hand, if the food was actually poisonous he will receive a negative reward (and also be reset). It is then up to the agent to remember where the food was and revisit the previous location. After the agent has eaten the non-poisonous food several times there is a random chance that the food and poison will be switched. The agent thus has to evolve to learn to detect from it's sensory input whether the food is poisonous or not, and can only know this by eating the food at least once to receive the associated negative or positive reward. 

The fitness of the agent is determined by the reward obtained in the environment, the better it is at remembering where food is and at avoiding eating poison, the higher the fitness. We also add an additional term that decays an agent's 'energy' every time step to encourage exploration. Eating food refreshes the energy again and gives the agent around 40 time steps to get new food. Once the agent runs out of energy it dies and will no longer be able to move or gather more rewards.  

To this end we provide several inputs to the agent, one that detects the wall of the maze, one that detects green and one that detects red. These inputs are passed to the first channel of the connected synapse ENU. We also have a neuron that provides the resulting reward of eating the food or poison. Positive rewards go to the second channel, while negative rewards to the third channel. For the output the agent can either go forward, left or right or do nothing (if no output neuron activates).  

The ENU network consists of 6 ENU neurons of which 3 are output neurons. Each neuron has 8 ENU synapses that sparsely connect to the other neurons (2 to the sensory neurons, 2 to the hidden neurons, 2 to the output neurons 1 to the reward neuron and 1 to itself). The neuron that has the highest output activity over 4 time steps determines the action of the agent (this allows for sufficient time of sensory input to propagate through the network). The output is taken from the first channel of the output neuron. We also add noise to the output gate of the ENU, since all ENUs share the same parameters and we want each neuron and synapse to behave slightly different upon initialization.

It is important to note that each generation the agents are reset and have no recollection of the past, each time they have to relearn which sensory neuron has what meaning and which output neuron performs what motor command. Also the network input and output neurons get randomly shuffled each generation, this avoids agents potentially exploiting the topology to learn fixed behaviour. We are thus evolving an ENU network to perform reinforcement type learning behaviour (evolving to learn), instead of directly learning fixed behaviour in the synaptic weights.

\paragraph{Experimental Details}
The experimental parameters were chosen through initial experimentation such that we achieved a good balance between computation time and performance. For optimization we used 1024 offspring and Gaussian mutations with standard deviation of 0.01, a learning rate of 1 and momentum of $0.9$. For both the synaptic and neuronal compartments an ENU with a memory size of 32 units was used, with 16 output units (i.e. 16 output channels). The experiments were run on a single Titan V GPU with an i7 12-core CPU. In case of the T-maze experiment the mean fitness was taken across 8 random environments for each offspring. For evolving the IAF and STDP model the mean over 32 environments was taken (allowing us to plot a STDP type curve from multiple observations). 1 episode in the environment is 1 generation and each episode the ENU internal memory states are reset. In case of the IAF and STDP environment each episode lasts for 100 time steps, while for the T-maze experiment the episode lasted for 400 time steps. 

\section{Results}

\begin{figure}[htb]
\includegraphics[width=0.49\columnwidth]{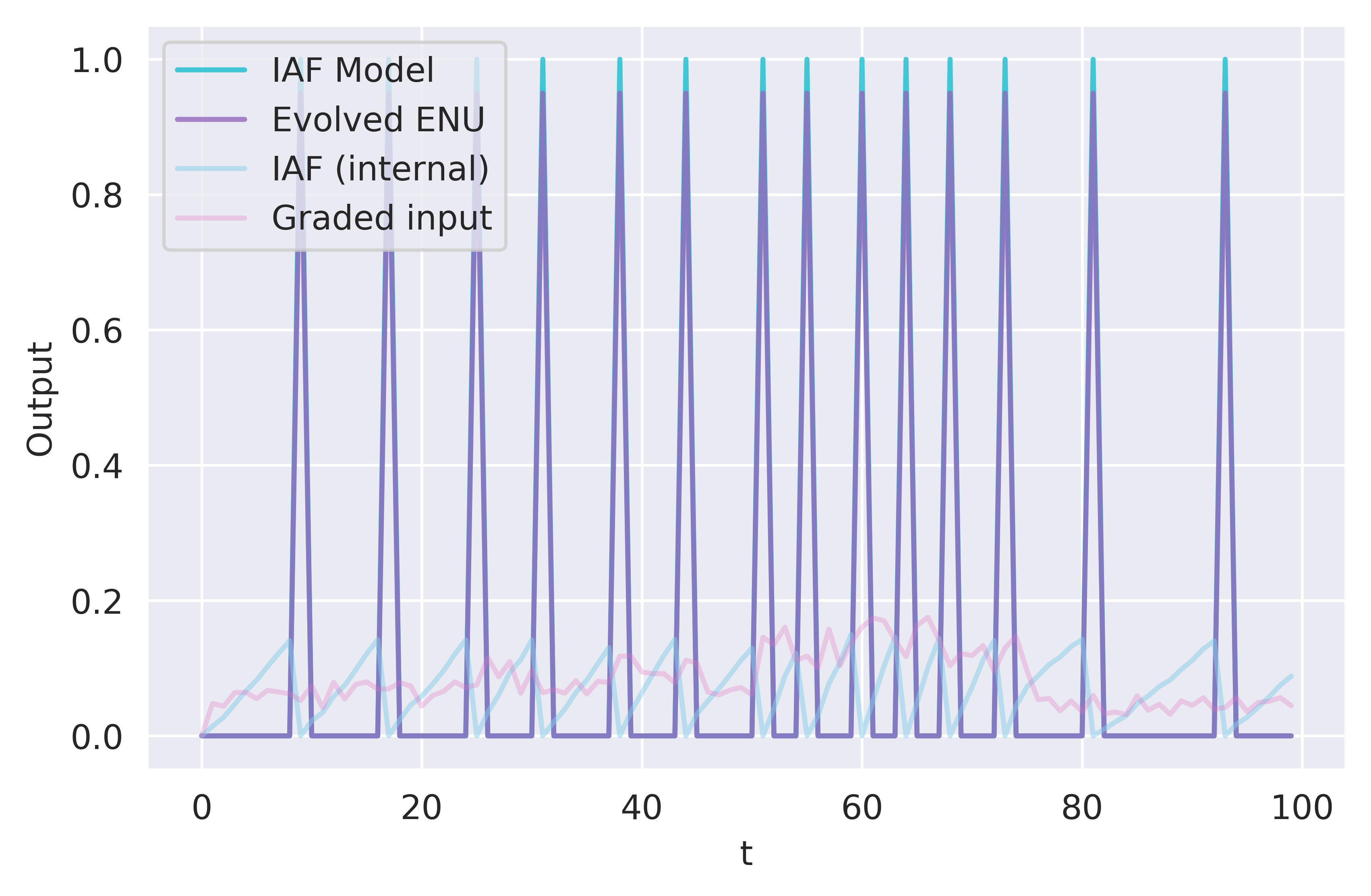}
\includegraphics[width=0.49\columnwidth]{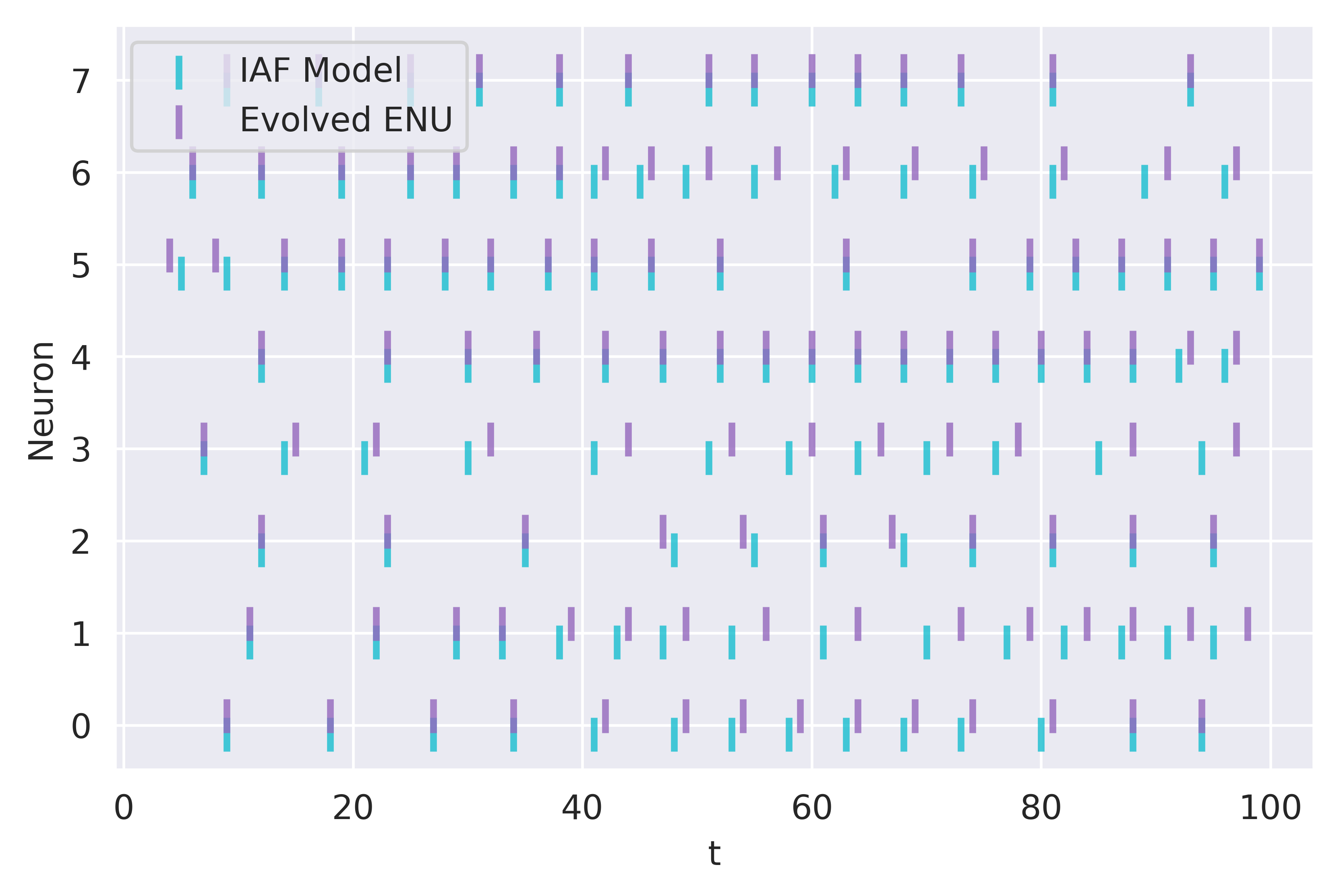}
\caption{\textbf{Result of evolving Integrate and Fire neurons}, for a single input example (left) and of multiple inputs (right). The evolved ENU model closely matches the actual IAF model, properly integrating information and spiking at the right time once a threshold is reached. It is also able to correctly process random graded input potentials that result in slower or faster spiking patterns. }
\label{results_iaf}
\end{figure}

\paragraph{Single ENU: Evolving Integrate and Fire Neurons}
Results of evolving Integrate and Fire neurons for 3000 generations are shown in figure \ref{results_iaf}. It can be seen that the ENU is able to accurately mimic the underlying IAF model, properly integrating the received graded input, and outputting a spike at the right time. This shows an ENU is capable of evolving to approximate spiking behaviour. 

\begin{figure}[htb]
\centering
\includegraphics[width=0.49\columnwidth]{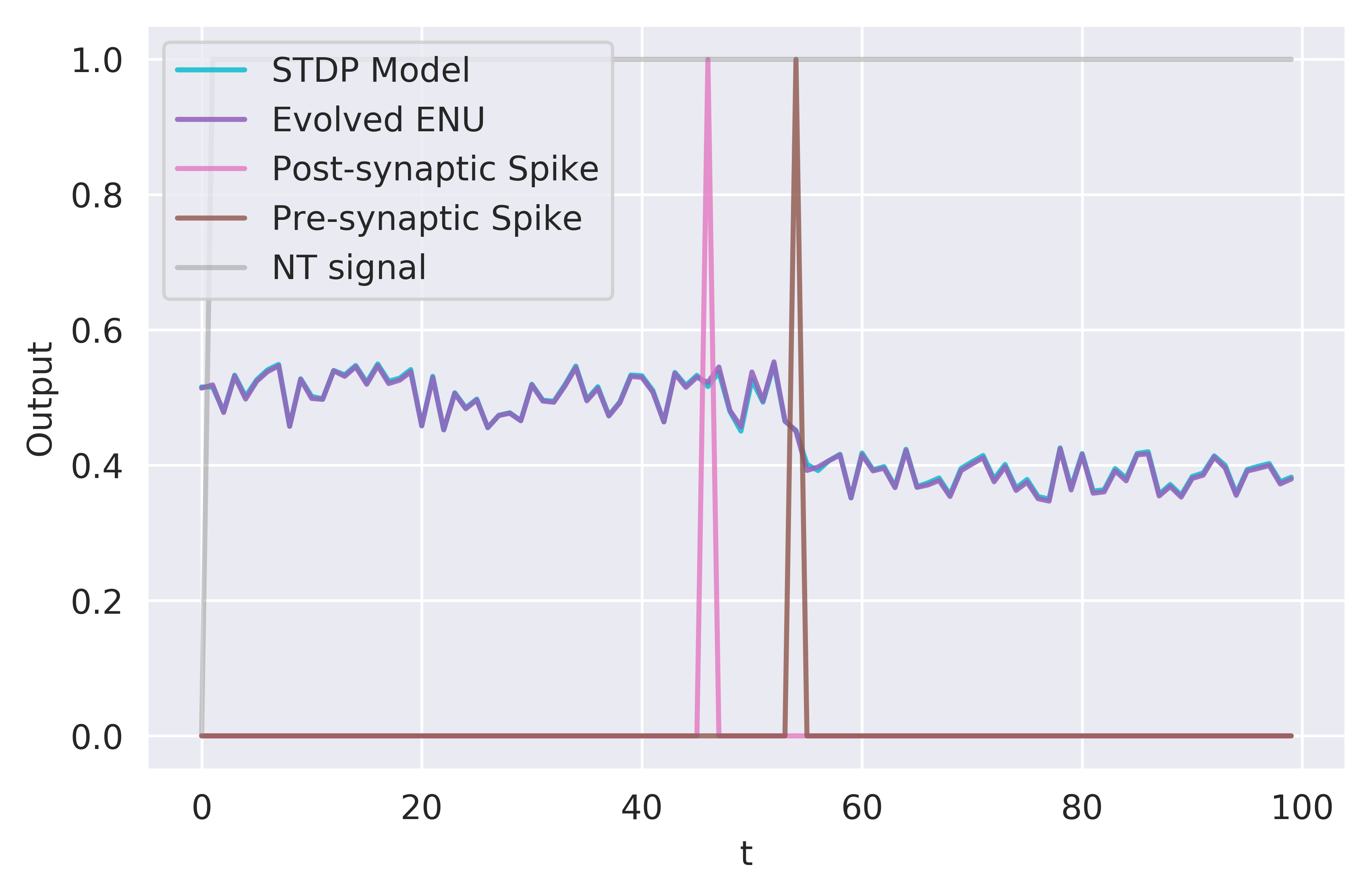}
\includegraphics[width=0.49\columnwidth]{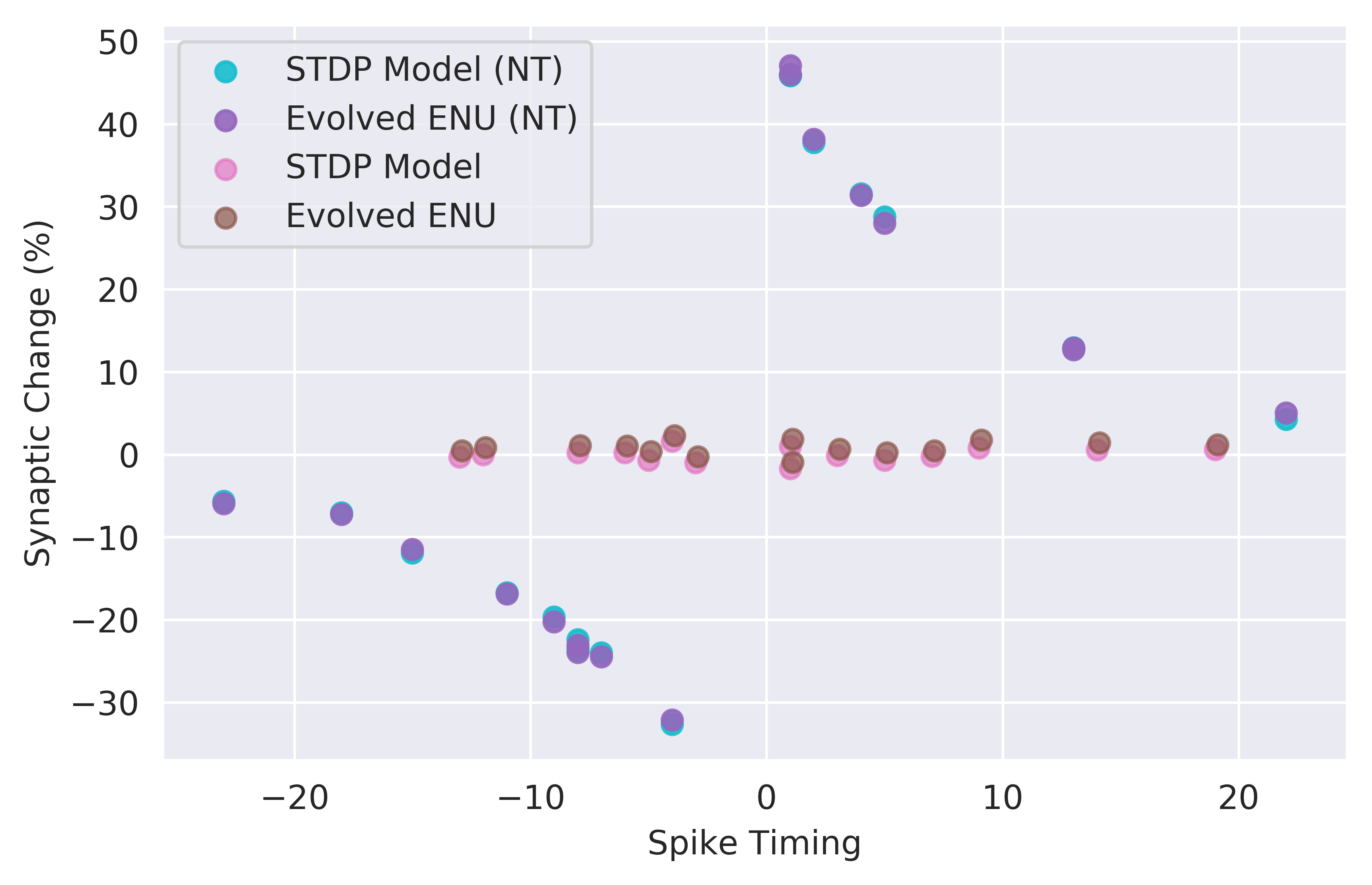}
\caption{\textbf{Results of evolving neuromodulated Spike-Timing-Dependent Plasticity}, for a single input example (left) and multiple input observations (right). It evolved to update the synaptic weight only when the neurotransmitter signal is present in the input, it also evolved to update the weights relative to the timing of the pre and post synaptic spike, matching the original STDP function.}
\label{results_stdp}
\vspace{-0.5cm}
\end{figure}

\paragraph{Single ENU: Evolving Neuromodulated STDP}
Figure \ref{results_stdp} illustrates the results after evolving an STDP type learning rule for 10000 generations. If the pre-synaptic spike occurs after the post-synaptic spike the graded input potential reduces in output intensity as in the standard STDP rule (and vice versa). This demonstrates an ENU is capable of evolving complex synaptic type learning through memorizing pre and post synaptic spikes in its internal memory and by storing and updating some dynamic parameter that changes how the incoming input is processed (analogous to a synaptic weight). It also had to evolve the multiplication operation of this dynamic parameter with the input, as initially the ENU has no such operation.  

\begin{figure}[!htb]
\centering
\includegraphics[width=0.7\columnwidth]{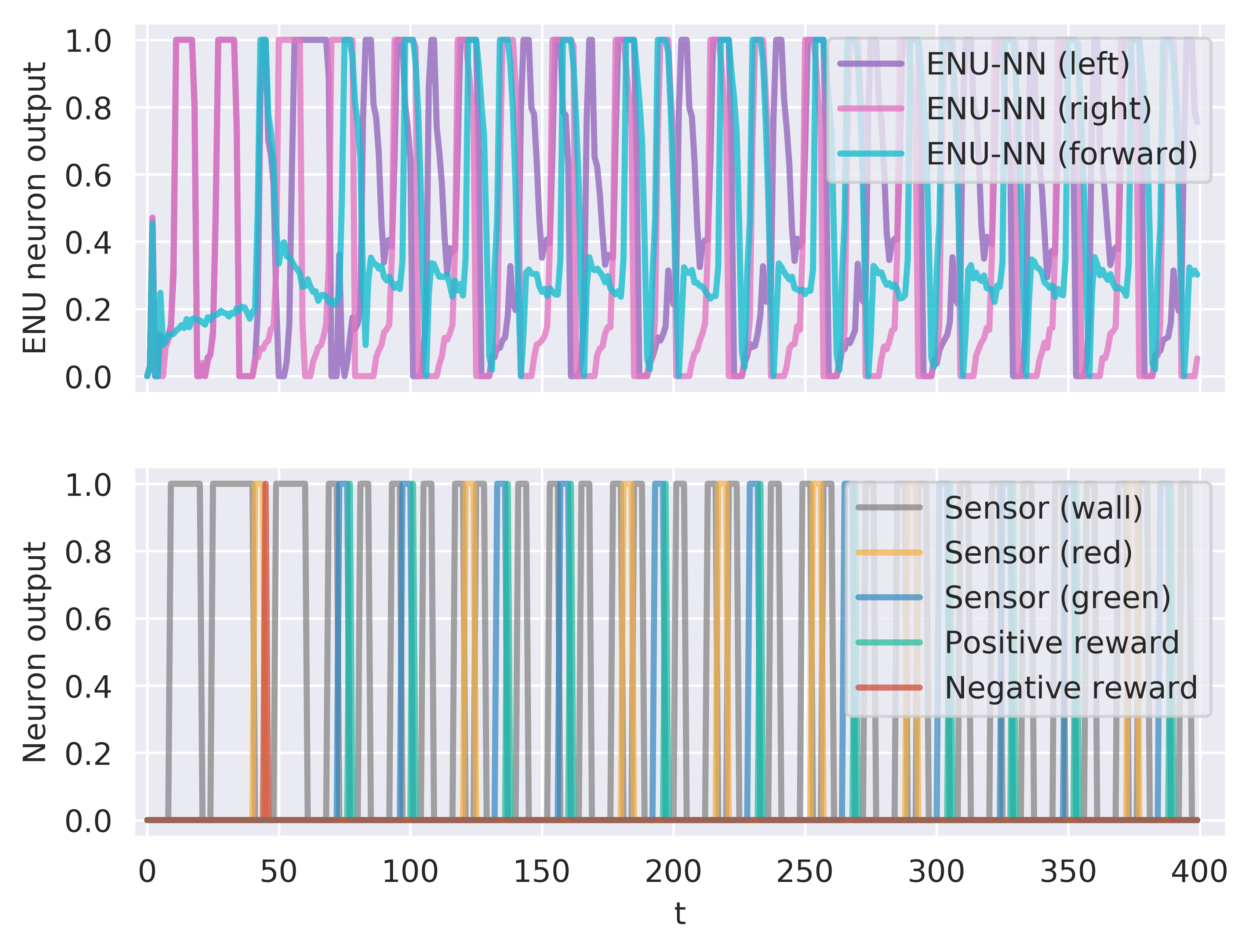}
\caption{\textbf{Results of evolving Reinforcement learning in a network of ENUs}, showing both the output of the agent (top) and input (bottom). Spike like patterns on the output neuron evolved completely independently, even though we did not directly optimize for such behaviour. Furthermore, it can be seen that the agent performs one-shot learning when eating the poison, only eating it once and subsequently always avoiding it.}
\label{results_network_output}
\end{figure}

\paragraph{Network of ENUs: Evolving Reinforcement Learning}
Results for evolving Reinforcement Learning behaviour in a network of ENUs after 30000 generations in a T-maze environment are shown in Figure \ref{results_network_output}. A single generation is one episode in the environment (one simulation run) and last for 400 time steps. Each generation all the dynamic parameters of each synapse and neuron reset, meaning it always has to relearn which sensory neuron leads to a negative or positive rewards and which output neuron performs what action, as if it is reborn (a blank slate).  

Interestingly we obtain spike like patterns on the output neurons even though we never strictly enforced it. We only optimize to maximize the reward in the environment. This could partially be explained by the sensory input neurons outputting spikes as well. However, the resulting output is not identical and it still had to evolve to process those input spikes and integrate them properly across the network and at the output. 

\begin{figure}[!hbt]
\centering
\includegraphics[trim={7cm 3cm 0.3cm 0},clip,width=0.85\columnwidth]{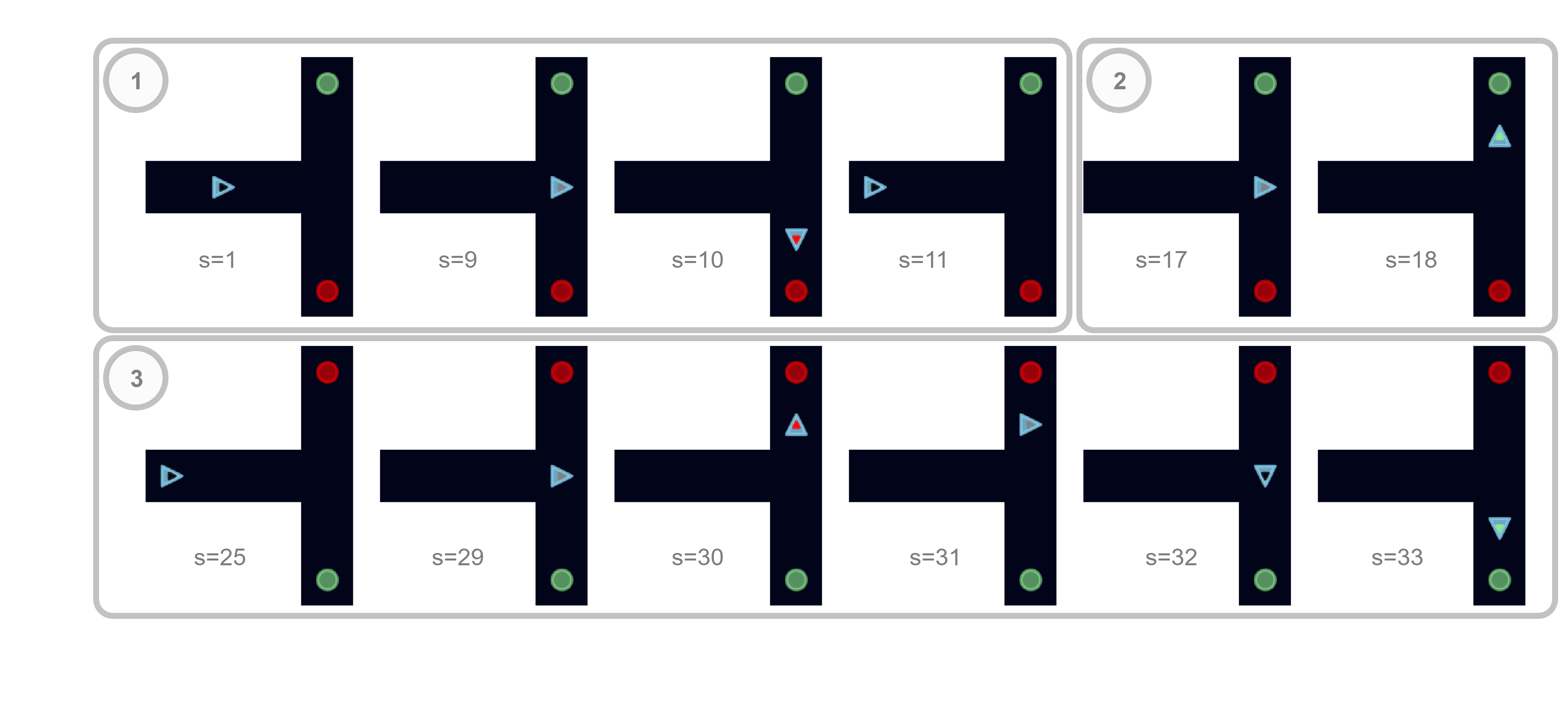}
\caption{\textbf{Example of the ENU Network evolved behaviour}. It evolved to learn to remember which food gives a negative reward (1), avoid such reward in the future by taking a different route (2), and adapt to turn around once the food and poison are switched (3).}
\label{results_network_rollout}
\end{figure}

An example of the steps taken by the agent are also given in figure \ref{results_network_rollout}. We can see that the agent at first detects red, eats the poison, and subsequently receives a negative reward. After that the agent learns to go the other way to get the food instead (detecting green). Once the food is switched it detects the poison but does not eat it (and so does not receive a negative reward), since it has now learned to turn around and obtain food on the other side instead. 

The performance (and fitness) is measured by how good the agent is at getting food and avoiding poison. The ENU networks of the agents with random mutations better able to learn and store the right information in the internal memory of each individual ENU will thus have a higher fitness, able to learn quicker from negative or positive rewards. Evolution Strategies then updates the shared parameters of the ENU gates in that direction. This leads it to improve its ability to process information in the ENU network and learn from reward feedback. It has to evolve in a way such that all ENU compartments with the shared gate parameters can cooperate and update their unique internal state to achieve an higher fitness. It thus evolves a type of universal function the same across all neurons and synapses that updates their internal dynamic parameters depending on the input.

The agent is able to learn through the evolved ENU network that the sensory neuron that detects the color red and the activation of the output neuron that results in eating it lead to a negative reward. This means that it evolved a mechanism for updating the ENU synapse between those neurons to have an inhibitory effect, such that next time it detects red, the previous action that led to eating the poison is not chosen (it is inhibited), and instead another action is taken.

\vspace*{-0.25cm}
\paragraph{Comparison to other models}
Standard Deep Neural Networks (DNNs) use only a single value for the synaptic weight parameter (which is static) and neuron output. The same holds for spiking neural network models with STDP type learning rules \cite{maass1997networks}. The activation function and update rules of these models are also fixed. Similarly, previous meta-learning approaches that learn to learn synaptic update rules still use the same fundamental underlying design of DNN computations \cite{andrychowicz2016learning, ha2016hypernetworks}. In our case, each dynamic parameter is a vector capable of storing multiple values that can influence the "operation" performed by the synaptic and neuronal ENU compartments. We evolve the ability of each ENU to update those dynamic parameters based on the reward obtained and local input received, unlike DNNs that use backpropagation over the entire network given some learning signal. In order for learning to occur in our ENU network, the synapse ENU has to evolve the performed operation on the input over multiple channels, which is not necessarily a multiplication as in standard DNNs.
\newpage 
Each generation our "dynamic" parameters also get reset, while in DNNs the parameters of the weights persist. DNNs are thus not learning-to-learn but learn fixed behaviour instead, only applicable to the current environment. Additionally, to learn from reward signals using deep neural networks, reinforcement based learning methods like Deep Q-learning \cite{mnih2015human} or Actor-Critic models \cite{mnih2016asynchronous} are required. In these methods however the agent requires millions of observations to update the actions to take in that specific environment. 

In our case the final evolved model is able to perform one-shot learning from a single experience to modify it's behaviour. This is because a learning rule evolves that is able to update the synaptic compartments given the reward, which consequently enables it to alter the way information flows through the synapses and neurons (over multiple channels). It also implicitly evolved random exploration like behaviour to seek out rewards, since this is required to achieve a higher overall fitness. Having multiple channels allows for more flexibility in learning and information processing behaviour, and could also explain why biological networks might have evolved to use so many types of different neurotransmitters \cite{lauder1993neurotransmitters, doya2002metalearning}.

\section{Discussion}
We showed that we were able to successfully train the proposed Evolvable Neural Units (ENUs) to mimic Integrate and Fire Neurons and Spike-Timing Dependent Plasticity. This demonstrated that in principle an ENU is flexible enough to evolve neural like dynamics and complex processing mechanisms. We then evolve an interconnected network where each synapse and neuron in the network is such an evolvable neural unit. This network of ENUs can be evolved to solve a T-maze environment task, which results in an agent capable of reinforcement type learning behaviour. This environment requires the agent to evolve the ability to learn to remember the color and location of food and poison depending on the reward obtained, change its behaviour to avoid such poison, and dynamically adapt to changes in the environment. These neuronal and synaptic ENUs in the network thus have to learn to work together cooperatively through local dynamics to maximize their overall fitness and survival, dynamically updating the way they integrate and process information.

Interesting future directions include simulating and evolving entire cortical columns by evolving smaller networks that are robust to changes in network size, allowing us to scale up the number of neurons and synaptic compartments after the evolutionary process. Furthermore, we could use separate ENUs to evolve e.g. dendrites, axons or the behaviour of different cell types seen across cortical layers. It is however still uncertain what aspects of biological neural networks are actually necessary or just byproducts of evolutionary processes. It also remains an open question what type of environment would be required in order for higher level intelligence to emerge, and research into open-ended evolution and artificial life might therefore be critical to achieve such intelligence.  

The Evolvable Neural Units presented in this paper offer a new direction for potentially more powerful and biologically realistic neural networks, and could ultimately not only lead to a greater understanding of neural processing, but also to an artificially intelligent system that can learn and act across a wide variety of domains.

\bibliographystyle{unsrt}
\bibliography{references}

\end{document}